%% file: ms.tex
\documentclass[10pt,a4paper]{article} 
\usepackage[hyperref]{tacl2018} 
\usepackage{times,latexsym}
\usepackage{url}
\usepackage[T1]{fontenc}
\usepackage{verbatim}
\taclfinalcopy
\setcitestyle{semicolon} 

\usepackage{xcolor}

\title{Context in Neural Machine Translation:\\A Review of Models and Evaluations}

\author{Andrei Popescu-Belis \\
  University of Applied Sciences of Western Switzerland (HES--SO)\\
  School of Management and Engineering Vaud (HEIG--VD)\\
  Route de Cheseaux 1, CP 521\\
  1401 Yverdon-les-Bains, Switzerland\\
  {\sf andrei.popescu-belis@heig-vd.ch}\\[10pt]
  \today}

\begin{document}
\maketitle
\begin{abstract}
This review paper discusses how \emph{context} has been used in neural machine translation (NMT) in the past two years (2017--2018).  Starting with a brief retrospect on the rapid evolution of NMT models, the paper then reviews studies that evaluate NMT output from various perspectives, with emphasis on those analyzing limitations of the translation of contextual phenomena.  \emph{In a subsequent version,} the paper will then present the main methods that were proposed to leverage context for improving translation quality, and distinguishes methods that aim to improve the translation of specific phenomena from those that consider a wider unstructured context.
\end{abstract}

\input{introduction}

\input{nmt-architectures}

\input{nmt-toolkits}

\input{general-nmt-eval}

\input{discourse-nmt-eval}

\input{context-nmt}



\section*{Acknowledgments}
The author is grateful to the Swiss National Science Foundation (SNSF) for support through the DOMAT project (n.\ 175693): On-demand Knowledge for Document-level Machine Translation.

\bibliographystyle{plainnat}
\bibliography{cnmt2018}
\end{document}

%% file: introduction.tex
\section{Looking Back on the Past Two Years} 

Neural network architectures have become mainstream for machine translation (MT) in the past three years (2016--2018).  This paradigm shift took a considerably shorter time than the previous one, which was from rule-based to phrase-based statistical MT models. Neural machine translation (NMT) was adopted thanks to its superior performance, and despite its higher computational cost (which has been mitigated by optimized hardware and software) or its need for very large training datasets (which has been addressed through back-translation of monolingual data and character-level translation as back-off).  The NMT revolution is apparent in the burst of the numbers of related scientific publications since 2017, as well as in the increased attention MT receives from the general media, often related to visible improvements in the quality of online MT systems.

While much remains to be done, especially for low-resource language pairs or for specific domains, the quality of the most favorable cases such as English-French or German-English news translation has reached unprecedented levels, leading to claims that it achieves human parity.  A remaining bottleneck, however, is the capacity to leverage contextual features when translating entire texts, especially when this is vital for correct translation.  Taking textual context into account\footnote{In this review, `context' refers to the sentences of a document being translated, and not to extra-textual context such as associated images.  Multimodal MT is an active research problem, but is outside our present scope.} means modeling long-range dependencies between words, phrases, or sentences, which are typically studied by linguistics under the topics of discourse and pragmatics.  When it comes to translation, the capacity to model context may improve certain translation decisions, e.g.\ by favoring a better lexical choice thanks to document-level topical information, or by constraining pronoun choice thanks to knowledge about antecedents.  

This review paper puts into perspective the significant amount of studies devoted in 2017 and 2018 to improve the use of context in NMT and measure these improvements.  We start with a brief recap of the mainstream neural models and toolkits that have revolutionized MT (Section~\ref{sec:nmt}). 
We then organize our perspective based on the observation that most MT studies design and implement models, run them on data, and apply evaluation metrics to obtain scores, i.e.\ Models + Data + Metrics = Results.  Novelty claims are generally made about one or more of the left-hand side terms, claiming improved results in comparison to previous ones.  

Existing MT models can be tested on new metrics and/or datasets, to highlight previously unobserved properties of these models.  Therefore, in Section~\ref{sec:eval}, we review evaluation studies of NMT, which either apply existing metrics (going beyond n-gram matching) or devise new ones.  We discuss these studies by increasing order of complexity of the evaluated aspects: first grammatical ones, and then semantic and discourse-level ones, including word sense disambiguation (WSD) and pronoun translation.  

Most often however, new models are tested on existing data and metrics, to enable controlled comparisons with competing models.  \emph{In an upcoming version of this paper,} we will discuss new NMT models that extend the context span considered during translation.  We will distinguish those that use unstructured text spans from those that perform structured analyses requiring context, in particular lexical disambiguation and anaphora resolution.


%% file: nmt-architectures.tex
\section{Neural MT Models and Toolkits}
\label{sec:nmt}

\subsection{Mainstream Models}
\label{sec:nmt-mainstream}



Early attempts to use neural networks in MT aimed to replace n-gram language models with neural network ones \citep{bengio2003neural,schwenk2006continuous}.  Later, feed-forward neural networks were used to enhance the phrase-based systems by rescoring the translation probability of phrases \citep{devlin2014fast}.  Variable length input was accommodated by using recurrent neural networks (RNNs), which offered a principled way to represent sequences thanks to hidden states.  One of the first ``continuous'' models, i.e.\ not using explicit memories of aligned phrases, was proposed by \citet{D13-1176}, with RNNs for the target language model, and a convolutional source sentence model (or a n-gram one).  To address the vanishing gradient problem with RNNs, long short-term memory (LSTM) units \citep{hochreiter1997long} were used in sequence-to-sequence models \citep{sutskever2014sequence}, and further simplified as gated recurrent units (GRU) \citep{D14-1179,chung2014empirical}.  Such units allowed the networks to capture longer-term dependencies between words thanks to specialized gates enabling them to remember vs.\ forget past inputs.

Such sequence-to-sequence models were applied to MT with an encoder and a decoder RNN \citep{D14-1179}, 
but had serious difficulties in representing long sentences as a single vector \citep{pouget2014overcoming}, although using bi-directional RNNs and concatenating their representations for each word could partly address this limitation.  The key innovation, however, was the attention mechanism introduced by \citet{bahdanau2015neural}, which allows the decoder to select at each step which part of the source sentence is more useful to consider for predicting the next word.\footnote{This paper was first posted on Arxiv in September 2014, while the one by \citet{D14-1179} was posted in June of the same year.}
Attention is a context vector -- a weighted sum over all hidden states of the encoder -- than can be seen as modeling the alignment between input and output positions.  The efficiency of the model was further improved, with small effects on translation quality \citep{D15-1166,D16-1137}.  The proposal for distinguishing local vs.\ global attention models by \citet{D15-1166} has yet to be incorporated in mainstream models.


The demonstration that NMT with attention-based encoder-decoder RNNs outperformed phrase-based SMT came at the 2016 news translation task of the WMT evaluations \citep{W16-2301}. The system presented by the University of Edinburgh \citep{W16-2323} obtained the highest ranking thanks particularly to two additional improvements of the generic model.  The first one was to use back-translation of monolingual target data from a state-of-the-art phrase-based SMT engine to increase the amount of parallel data available for training \citep{P16-1009}.  The second one was to use byte-pair encoding, allowing translation of character n-grams and thus overcoming the limited vocabulary of the encoder and decoder embeddings \citep{P16-1162}.  Low-level linguistic labels were shown to bring small additional benefits to translation quality \citep{W16-2209}.  The Edinburgh system was soon afterward open-sourced under the name of Nematus \citep{JunczysDowmunt2016}.


Research and commercial MT systems alike were quick to adopt NMT, starting with the best-resourced language pairs, such as English vs.\ other European languages and Chinese.  Around the end of 2016, online MT offered by Bing, DeepL, Google or Systran was powered by deeper and deeper RNNs (as far as information is available).  In the case of DeepL, although little information about the systems is published, its visible quality\footnote{See \url{https://www.deepl.com/en/quality.html} and the beginning of Section~\ref{sec:eval-bleu-grammar} for an estimate.} could be partly explained by the use of the high-quality Linguee parallel data.  

An interesting development have been the claims for ``bridging the gap between human and machine translation'' from the Google NMT team in September 2016 on EN/FR and EN/DE \citep{GoogleNMT}, and for ``achieving human parity on $\ldots$ news translation'' from the Microsoft NMT team in March 2018 on EN/ZH  \citep{Hassan2018AchievingHumanParity}.  These claims have raised attention from the media, but have also been disputed by deeper evaluations (see Section~\ref{sec:eval-bleu-grammar}).


RNNs with attention allow top performance to be reached, but at the price of a large computational cost.  For instance, the largest Google NMT system from 2016 \citep{GoogleNMT}, with its 8 encoder and decoder layers of 1,024 LSTM nodes each, required training on 96 nVidia K80 GPUs for 6 days, in spite of massive parallelization (e.g.\ running each layer on a separate GPU).  A more promising approach to decrease computational complexity is the use of convolutional neural networks for sequence to sequence modeling, as proposed by \citet{pmlr-v70-gehring17a} in the ConvS2S model from Facebook AI Research.  This model outperformed \citeauthor{GoogleNMT}'s system on WMT 2014 EN/DE and EN/FR translation ``at an order of magnitude faster speed, both on GPU and CPU''.  Posted in May 2017, the model was outperformed the next month by the Transformer \citep{NIPS2017_7181}.

The Transformer NMT model \citep{NIPS2017_7181} removes sequential dependencies (recurrence) in the encoder and decoder networks, as well as the need for convolutions, and makes use of self-attention networks for positional encoding.  For instance, the encoder is composed of six pairs of 512-dimensional layers; in each pair, the first layer implements multi-head self-attention, while the second is a fully-connected feed-forward layer. In the decoder, an additional layer in each pair implements multi-head attention over the encoder's output.  As a result, training on GPUs can be fully parallelized, thus substantially reducing training time, and slightly outperforms RNN models. 

For these reasons, the Transformer was quickly adopted by the research community: it was used by virtually all systems for the WMT 2018 news task \citep{W18-6401}.  The model is now implemented in most NMT toolkits (see next section).  While the Transformer remains the state-of-the-art at the time of writing, several of its authors have shown that RNN architectures could be improved beyond the Transformer using some of its insights, and that hybrid architectures based on RNN, CNN and the Transformer pushed the scores on WMT'14 EN/DE and EN/FR datasets even higher \citep{P18-1008}.\footnote{Furthermore, inspiration from the Transformer can be even found in the BERT language modeling technique (Bidirectional Encoder Representations from Transformers), also from Google, which reached new state-of-the-art results on 11 NLP benchmarks, including question answering or inference \citep{devlin2018bert}.}  A deeper attention model for MT has been presented at the end of 2018, filtering attention from lower to higher levels over five layers \citep{zhang2018DeepAttention}, with encouraging results.

The findings of the WMT 2018 news translation task \citep{W18-6401} confirmed the merits of the Transformer, though certain improvements in the architecture allowed a late-coming submission from Facebook \citep{W18-6301} to be ranked first on EN/DE.  This system trained the Transformer using reduced numeric precision, thus accelerating ``training by nearly 5x on a single 8-GPU machine''.  Results were further improved by the team using advances in back-translation to generate synthetic source sentences \citep{D18-1045}, with training sets reaching hundreds of millions of sentences; the system also achieved state-of-the-art performance on WMT '14 EN/DE test sets.

%% file: nmt-toolkits.tex
\subsection{NMT Toolkits}
\label{sec:nmt-toolkits}

The above NMT models are often available as open-source implementations in MT toolkits, which are built upon general-purpose machine learning frameworks supporting neural networks.  Machine learning frameworks are evolving at a rapid pace and so do NMT implementations.\footnote{Among the most recent changes in the ML ecosystem, one can cite the merger of Caffe into PyTorch, the growth of PyTorch itself in comparison to the earlier LuaTorch, the end of Theano development by the University of Montreal, and the integration of Keras into the core of TensorFlow.}  Most  NMT toolkits are now built over TensorFlow and Torch (Lua or Python), though others are built from scratch or over other frameworks.


The NMT toolkits most frequently used in research studies, including submissions to shared tasks, are the following ones (in alphabetical order).
\begin{itemize} \setlength{\itemsep}{0pt}
	\item \textit{DL4MT}: Deep Learning for Machine Translation. \url{https://github.com/nyu-dl/dl4mt-tutorial}.
	\item \textit{FairSeq}: Facebook AI Research Sequence-to-Sequence Toolkit.  \url{https://github.com/pytorch/fairseq}.
	\item \textit{Marian}: Fast Neural Machine Translation in C++ \citep{P18-4020}.  \url{https://marian-nmt.github.io/}.
	\item \textit{Nematus}: Open-Source Neural Machine Translation in TensorFlow \citep{E17-3017}.  \url{https://github.com/EdinburghNLP/nematus}.
	\item \textit{Neural Monkey} \citep{helcl2017neural}. \url{https://github.com/ufal/neuralmonkey}.
	\item \textit{OpenNMT} \citep{P17-4012}. \url{http://opennmt.net/}. 
	\item \textit{Sockeye} \citep{hieber2017sockeye} \url{https://github.com/awslabs/sockeye}.
	\item \textit{Tensor2Tensor} \citep{W18-1819}. \url{https://github.com/tensorflow/tensor2tensor}.
\end{itemize}

%% file: general-nmt-eval.tex
\section{How Good is NMT?  Fine-grained Evaluation Studies}  
\label{sec:eval}

Most proponents of novel NMT models evaluate them using the BLEU metric \citep{P02-1040} on parallel data from the WMT conferences (e.g.\ \url{http://statmt.org/wmt18}).  While this is a fairly common and accepted procedure, the significance\footnote{`Significance' meaning here `importance' or `relevance', and not statistical significance, which is often duly tested \citep{W04-3250}.} of small increases in BLEU is not entirely clear, especially in terms of perceived quality, given the multiplicity of quality aspects that are actually relevant to users \citep{hovy2002principles}.  
Human rating of quality, e.g.\ through direct assessment \citep{graham2017can}, is generally carried out only yearly, at dedicated evaluation campaigns such as WMT or IWSLT, often without delving into specific quality attributes any further.

Therefore, a rich set of evaluation studies have attempted to shed light on the various improvements brought by NMT, often compared to SMT.  These studies applied existing metrics, or devised new ones, using new or existing data sets, to assess fine-grained quality aspects of NMT output from various systems.  

The studies presented in this section evaluate existing NMT systems, without propose new techniques addressing the observed limitations (such proposals are discussed in Section~\ref{sec:dlnmt} below)).  We organize this section according to the linguistic complexity of the quality aspects (or attributes) of NMT output, from words to texts.
After a preliminary discussion of two studies using BLEU in various conditions, we consider evaluations of morphology, the lexicon, verb phrases, and word order (Section~\ref{sec:eval-bleu-grammar}).  When then turn to evaluations of contextual factors, from semantic properties including word sense disambiguation and lexical choice~(\ref{sec:eval-wsd}), to discourse-related or document-level quality aspects, in particular the translation of pronouns~(\ref{sec:eval-pronouns}).\footnote{Studies of other system qualities such as efficiency, adaptability, or usability have been comparatively less frequent and are not included here.  Properties such as the ability to handle multilingual -- as opposed to bilingual -- models and to perform zero-shot translation have been examined, e.g.\ by \citet{C18-1054}.)}

\subsection{Studies Using BLEU}

Claims about performance can be thoroughly analyzed even when BLEU is used as a metric.  Recently, \citet{Toral2018Attaining} examined again the claim for human parity on EN/ZH translation from the Microsoft NMT team \citep{Hassan2018AchievingHumanParity} by inspecting more closely the test sets.  One finding is that a significant portion of the data was originally written in English, so the system's Chinese source was ``translationese'', i.e.\ influenced by the target language.\footnote{This is additional empirical evidence for the need of properly constructed \emph{directional corpora}, e.g.\ extracted from Europarl with additional speaker information \citep{CartoniMeyer2012}.}  If evaluation is restricted to original English source sentences, then human parity is not reached.  Moreover, many reference translations have fluency problems, contain grammatical errors or mistranslated nouns.  The authors also confirm the finding of \citet{D18-1512} that professional human judges, who also have higher inter-rater agreement, still find a gap between human translations and NMT.

\citet{W17-3204} analyze evaluation results of NMT and PBSMT with the BLEU metric and make several observations: the quality of NMT decreases quickly out of the training domain, and with long sentences; the attention model is not necessarily a true alignment model, and the beam search leads to acceptable results for narrow beams only.  The authors infer six challenges for NMT, but discourse-level metrics point to additional challenges, in particular related to the use of context (see Section~\ref{sec:eval-discourse}).

\subsection{Grammatical and Lexical Qualities of NMT}
\label{sec:eval-bleu-grammar} 

Shortly after the NMT approach became state-of-the-art, several finer-grained evaluations than those based on BLEU were applied to it.  These studies differ widely in the granularity of error classifications, and in how error metrics are applied and on what data, as we now discuss.

\subsubsection{Human and Automatic Ratings}
%
%
One of the first detailed analyses of the output of NMT (encoder-decoder RNNs with attention) in comparison with SMT was presented by \citet{D16-1025}.  Using high-quality post-edits by professional translators on system outputs from the IWSLT EN/DE 2015 task, errors were automatically detected and classified according to several categories: morphological errors (correct lemma but wrong form), lexical errors (wrong lemma), and word order errors.  The latter type was further analyzed according to POS and dependency tags; but lexical errors were not further subcategorized (see Section~\ref{sec:eval-wsd} for such attempts).

The comparisons showed that NMT made about 20\% fewer lexical or morphological mistakes than SMT, and up to 50\% fewer word-order errors (especially on verb placement, which is essential in German), thus demonstrating better flexibility than SMT for language pairs with different word orders.  However, NMT sometimes failed to translate all source words, such as negations, which is detrimental to adequacy and difficult to spot by the user.  The authors also found that the Translation Error Rate (TER) of all systems increased similarly with sentence length, and NMT outperformed PBSMT, though by a smaller margin on longer sentences.  An extended version of the study \citep{BENTIVOGLI201852} added an analysis of IWSLT EN/FR data which confirmed the above conclusions, and found that NMT had a better capacity to reorder nouns in EN/FR translation than PBSMT, but made more errors on proper nouns.

\citet{Popovic2017Comparing} performed error analysis on 267 EN/DE and 204 DE/EN sentences from WMT 2016 News Test, and compared the output of an NMT system \citep{W16-2323} with one from PBSMT (based on Moses), both obtained from WMT submissions.  A variety of grammatical aspects were evaluated, showing that morphology (particularly word forms), English noun collocations, word order, and fluency are better for NMT than PBSMT.  Still, the tested RNN-based NMT system had difficulties with polysemous English source words, and with English continuous verb tenses (on the target side).  In an extended version, \citet{Popovic2018Language} confirmed these conclusions, and added analyses of English-Serbian translation on 267 sentences.


In an early comparison of PBSMT to NMT, \citet{castilho2017comparative} required professional translators to post-edit MT output, namely 100 English sentences from MOOCs, translated into German, Portuguese, Russian and Greek.  Translators ranked outputs from the two systems, and counted the time and number of operations used for post-editing; fluency and adequacy were also rated.  Specific error annotation was performed as well, dividing errors into several classes: inflectional morphology, word order, and omission~/ addition~/ mistranslation.  NMT globally outperformed PBSMT on these metrics, except for omission and mistranslation.  It also outperformed PBSMT on post-editing time, as NMT errors were more difficult to grasp, although fewer sentences needed correction.  These findings were confirmed in a subsequent article \citep{castilho2017neural} which added two additional use cases beyond MOOCs: EN/DE translation of product ads, and ZH/EN patent translation.  NMT thus appeared superior in fluency, but superiority in adequacy or post-editing effort was not observed.  The use of NMT as an assistance tool for professional translators appeared as uncertain.\footnote{In the meanwhile, the switch of virtually all online MT offerings to NMT tends to indicate a consensus on the advantages of NMT for web translation.}

Given the multiplicity of translation directions and domains that can be tested, it may be of no surprise that other studies followed suit.  
\citet{toral-sanchezcartagena:2017:EACLlong} evaluated PBSMT and NMT submissions to WMT 2016 for 9 translation directions (EN to/from CS, DE, FI, RO, RU, except FI/EN) and confirmed that NMT is more fluent (measured with an edit distance) and has better inflected forms, but struggles with sentences longer than 40 words.
In a larger journal article submitted in August 2017, \citet{Klubicka2018Quantitative} apply a multidimensional quality metric (MQM) and study the statistical significance of differences between MT systems, for English to Croatian, a morphologically rich language.  MQM is applied by two human raters over 100 sentences, with outputs from 3 systems, with a large taxonomy of error types, such as word order, agreement, spelling, along with omission and mistranslation.  The authors found that their best system (NMT Nematus) reduced the error of their weakest one (PBSMT Moses) by about 50\%, and was especially better for long-distance grammatical agreement. 

\citet{Burchardt2017Comparing} created a large test suite of around 5000 EN/DE segments to evaluate MT output for 120 phenomena grouped in 15 categories (e.g. `ambiguity', `function words', or `long-distance dependency').  They used about 800 items for a comparison of 7 MT systems (rule-based, PBSMT, or neural), and reached somehow surprising conclusions, likely because scores were micro-averaged across categories of test sentences with very different sizes (e.g.\ 529 out of 777 test verb tense~/ aspect~/ mood).  Had macro-averaging been used, the likely winner would have been the Google NMT system \citep{GoogleNMT}, which performed best on most error categories.  

Another error taxonomy was proposed by \citet{Esperanca2017Evaluation} and was applied to PBSMT and NMT output over the BTEC-corpus, to compare reference-based metrics with explicit error annotation and study the translators' perception of the output.  Again, NMT outperformed PBSMT, albeit slightly.  Similarly, \citet{VANBRUSSEL18.611} compared the outputs of online systems on EN/NL translation.  NMT was found to be particularly fluent, although omissions remained a problem, and made fewer WSD errors but more mistranslations, which may be harder to post-edit.


\subsubsection{Contrastive Pairs and Challenge Sets}

Evaluation methods based on 
\emph{contrastive pairs} require access to the probability estimates of pairs of source and target sentences from the evaluated system.  These probabilities are easy to obtain from NMT systems that are not used as black boxes, but impossible to get from online systems.  Moreover, these methods do not guarantee that if a systems correctly scores two candidate target sentences, then it can also find the correct translation using beam search when only the source is given.


\citet{E17-2060} designed LingEval97, a test set of 97k contrastive pairs, built from reference EN/DE translations from WMT.  A reference translation can be modified in five different ways to generate an incorrect counterpart, using editing rules to automate the process for a large set.  Incorrect sentences are generated by (1) changing the gender of a singular determiner; (2) changing the number of a verb; (3) changing a verb particle; (4) changing aspects of sentence polarity, e.g.\ inserting or deleting a negation particle; (5) swapping characters in unseen names.  The main findings are that character-based NMT systems (RNN-based) are better than byte-pair encoding ones on type~5 errors, but worse on types~1, 2 and~3.  As for polarity, while spurious insertions of negations are well detected by all studied systems, 
spurious deletions are less well detected, echoing he fact that negations are sometimes omitted in NMT output.

LingEval97 was recently reused in a comparison of RNN, CNN and Transformer architectures by \citet{D18-1458}, along with ContraWSD set presented below for a semantic evaluation.  Performance on LingEval97 appeared to be quite similar across architectures, with RNNs being particularly competitive for modeling long-distance agreement between subjects and verbs (in fact, detecting wrong agreements).

\citet{D17-1263} proposed a linguistically-motivated test suite or more exactly a challenge set, as the sentences are not accompanied by a reference translation -- instead, human judges are required to evaluate whether each challenge sentence was translated correctly or not.  The application cost remains moderate due to the small amount of sentences: 108 for EN/FR translation.  The sentences are divided into three categories: morpho-syntactic (including agreement and subjunctive mood), lexico-syntactic (including double-object and manner-of-movement verbs), and syntactic (e.g.\ yes-no and tag questions, and placement of clitic pronouns).  At the end of 2016, the challenge set was applied to PBSMT and NMT \citep{W16-2323,GoogleNMT}.  Later on, it was also applied to the online DeepL Translator\footnote{\url{https://medium.com/@pisabell/}} showing a 50\% error reduction with respect to the best NMT system of the initial article.

%% file: discourse-nmt-eval.tex
\subsection{Evaluation of Semantic and Discourse Phenomena in NMT Output}
\label{sec:eval-discourse}

Categorizing MT errors as `semantic' or `discourse' is not always clear-cut, as it often involves an hypothesis on the cause of an error.  For instance, is outputting a wrong pronoun a morpho-syntactic or a discourse error?  If only its gender is wrong, then this may be attributed to ignorance of its antecedent, whereas if the case is wrong (subject vs.\ object), then the error can be considered as grammatical.  
In this section, we present NMT evaluation studies focusing on errors that can be attributed to insufficient knowledge or modeling of semantic and discourse properties, which often require considering a context made of multiple sentences.\footnote{This contrasts with the local view of context adopted e.g.\ by \citet{D18-1339}, where context actually means the left bigram context.}  

We group studies into three categories: those dealing with lexical choice (including WSD and lexical coherence), those dealing with referential phenomena (anaphora and coreference), and finally those dealing with discourse structure, though no study among the latter deals with NMT.

\subsubsection{Evaluation of Lexical Choice: WSD and non-WSD Errors}
\label{sec:eval-wsd}

\paragraph{Notations.} 
Word ambiguity is often cited as an obvious difficulty for translation.  In reality, ``ambiguity'' is a complex notion, and we will focus in this section on content words (open class).  Let us suppose ideally that a word $w$ may convey one or more language-independent senses $s_1, s_2, \ldots,$ as listed for instance in WordNet, and that a given occurrence of $w$ conveys only one sense at a time.  Let us now consider independently a word $w^f$ in the source language, and three words $w^e_1$ to $w^e_3$ in the target language, with the following senses: $w^f: \{s_1, s_2, s_3\}$, $w^e_1: \{s_1, s_2\}$, $w^e_2: \{s_2, s_3\}$, $w^e_3: \{s_3, s_5\}$.

All these words except $w^e_2$ and $w^e_3$ can be said to be ambiguous, as they may convey more than one sense.  However, for translation, only the ambiguity of $w^f$ actually matters.  If an occurrence of $w^f$ conveys sense $s_1$ but is translated with $w^e_2$ (which cannot convey this sense), this is called a \emph{word sense disambiguation (WSD) error}, regardless of how $w^e_2$ was actually found, i.e.\ whether or not WSD was explicitly performed on the source side.  If, however, the occurrence of $w^f$ conveys sense $s_2$, then both $w^e_1$ and $w^e_2$ can be used, in principle.  Then, regardless of what a reference translation may contain, using one of $w^e_1$ or $w^e_2$ cannot be a WSD error.\footnote{Note that if a word $w'^f$ may convey only one sense, there is no potential for a source-side WSD error.  Note also that translating a word $w'^f$ by a word that can convey all its senses does not oblige the system to perform source-side WSD.} 

This representation does not account for the additional constraints that may distinguish between translations by $w^e_1$ and $w^e_2$ when $w^f$ conveys sense $s_2$, and which may lead to \emph{non-WSD lexical errors}.  For instance, it may happen that $s_2$ is an infrequent sense of $w^e_1$, or if they are verbs, $w^e_1$ and $w^e_2$ may have different sub-categorization frames.  If a previous occurrence of $w^f$ was translated by $w^e_2$, it may be the case that word repetition is necessary for cohesion, or for understanding a repeated reference, ruling out a subsequent translation by $w^e_1$.  Other constraints may come from collocations (MWEs) or terminology.  Some of these factors are mere preferences, while others are strong constraints, leading to genuine mistakes if not respected.  Therefore, non-WSD lexical errors may violate cohesion, coherence, sense frequency distributions, collocations, terminology, or grammatical constraints.

For instance, the test set designed by \citet{N18-1118} for ambiguous source words (like $w^f$ above) equates WSD errors to coherence errors, because they generally make the output incoherent.  Conversely, non-WSD errors are equated to cohesion errors, although the authors concede that ``these types are not mutually exclusive and the distinction is not always so clear.''  While cohesion errors \emph{per se} can in principle be counted automatically \citep{D12-1097}, WSD errors, as well as non-WSD errors not related to cohesion (e.g.\ due to collocations or terminology) are more difficult to spot without human intervention.  One solution is the recent trend -- though not without remote ancestors \citep{C90-2037} -- to use contrastive pairs containing ambiguous source words, which we discuss hereafter.

\paragraph{Test suites with contrastive pairs.} 
Based on the same principle as Lingeval97 \citep{E17-2060} mentioned above, ContraWSD is a set of contrastive pairs intended to evaluate the capacity of an MT system to translate the correct sense of a polysemous word in context \citep{W17-4702}.  About 80 word senses were selected automatically, by observing target-side variation, for each of the DE/EN and FR/EN pairs.\footnote{See \url{https://github.com/a-rios/ContraWSD}.}  For each sense, 90 sentences are available on average, and for each reference translation, an average of 3.5 and 2.2 wrong translations are generated by replacing the target word with other observed translations of the word.  As with Lingeval97, a system can be tested with ContraWSD only if it can output 
the probability of a source/target sentence pair, which excludes black box systems, and a good answer is a case where the system ranks a correct translation higher than an incorrect one, given the source.  The authors found that a baseline NMT system \citep[Nematus,][]{E17-3017} reached about 70\% accuracy, compared to 93--95\% for a human.  The sense-aware systems proposed in their study remained in the same accuracy range on average (see below), but scored higher when disambiguating frequent words.

This approach was pursued and proposed as a supplementary test suite at WMT18 \citep{W18-6437}, where it was formulated as a classic translation task, with a test set of 3,249 DE/EN sentence pairs (from several corpora on OPUS) which contained ambiguous German words identified in ContraWSD.  The scoring is automatic in most cases, by observing the presence of a known correct vs.\ incorrect translation of each polysemous source word.  All systems submitted to the WMT18 news translation task \citep{W18-6401} were also evaluated for WSD, and compared to certain 2016 systems, finding that accuracy of the best system progressed from 81\% to 93\%, and that the correlation with BLEU scores was strong but not perfect. 

Another contrastive test set was made available in November 2017 by \citet[Section 2.1]{N18-1118} to assess lexical choice in English/French translation, but also pronoun choice (see next section).  The set is thus composed of two equally sized subsets, each consisting of `blocks' based on modified movie subtitles.  There are 100 blocks testing lexical choice capabilities (WSD and non-WSD) \citep[see also][Section~7.1]{Bawden2018Thesis}. 
Formally, let us denote a block as $((C^f_1, S^f, C^e_1, S^e_\alpha, S^e_\beta), (C^f_2, S^f, C^e_2, S^e_\beta, S^e_\alpha))$.
Each block is based on a source sentence $S^f$ containing a polysemous word $w^f$.  Two different source sentences $C^f_1$ and $C^f_2$ are provided as context, i.e.\ preceding $S^f$.  Their role is to modify the sense conveyed by the occurrence of $w^f$ in sentence $S^f$.  For each context, the block provides a correct translation of $S^f$ ($S^e_\alpha$ in the first case, $S^e_\beta$ in the second case), along with an incorrect one ($S^e_\beta$ and respectively $S^e_\alpha$).  The reference translations of the context sentences ($C^e_1$ and $C^e_2$) are also included.
Because the source sentence is kept constant for the two contexts, a non-contextual system would provide the same answer for both contexts (i.e.\ same ranking of true/false candidates) and obtain 50\% accuracy. 
Among the 100 blocks provided by \citet{N18-1118}, some are designed to test WSD capabilities, and include contexts such as $C^f_1$ indicates that $w^f$ conveys sense $s_1$ (with the notations above), so the correct translation is $w^e_1$ and the incorrect one is $w^e_2$. Then, context $C^f_2$ indicates that $w^f$ conveys sense $s_3$, and reverses the correctness of $w^e_1 / w^e_2$ translations.  Other blocks test non-WSD related lexical choices, which rely more on the target contexts $C^e_1$ and $C^e_2$ for deciding which translation is correct, e.g.\ the need to repeat the same word.


\paragraph{Exploring attention to context for NMT of polysemous words.} 
A quantitative analysis of the WSD capacities of NMT (encoder-decoder RNN with attention) was provided by \citet*{N18-1121} in August 2017, who opted for straightforward criteria to identify polysemous words and assess their translations.  The number of senses of each EN source word (for EN/DE, EN/FR and EN/ZH NMT) was extracted from the online \textit{Cambridge Dictionary}, and correct translation meant identity to the reference.  Further on, to demonstrate the benefits of their proposed NMT improvements (see below), they restricted the list of polysemous words to a list of 171 English homographs found on Wikipedia.

The findings presented by \citet{W18-1812} may explain why the capabilities of baseline NMT systems (RNN-based built with OpenNMT-py for EN/FR translation over Europarl and NewsComments) for WSD remain quite limited.  They examined the representations of occurrences of polysemous words at various levels of the NMT encoding layers.  Specifically, the tests involved 426 sentences and four polysemous words (\emph{right}, \emph{like}, \emph{last}, and \emph{case}), and showed that the encoded context seems insufficient to enable WSD in most cases.

This is confirmed by \citet{W18-6304} who directly looked at how attention is distributed when translating polysemous words from ContraWSD.  They compared RNN encoder-decoder with the Transformer model, with two ways to compute translation accuracy on polysemous words: either by comparing directly with a word-aligned reference, or by scoring the contrastive pairs as in \citep{W17-4702}.  In both cases, the Transformer clearly outperforms the RNN, though performance appears to be lower with reference-based scoring.
The main findings are that attention weights are more concentrated on the ``ambiguous noun itself rather than context tokens'' and that ``attention is not the  main  mechanism used by NMT models to incorporate contextual information for WSD.''

ContraWSD was again put to use by \citet{D18-1458} for a quantitative evaluation of WSD for DE/EN and DE/FR translation. 
The comparison of RNNs, CNNs and Transformer showed that the latter is significantly better than the other two, likely because the network ``connects distant words via shorter network paths'', but no further explanation or analysis on WSD is provided.


\subsubsection{Evaluation of Pronouns and Coreference}
\label{sec:eval-pronouns}

A revival of the interest in improving discourse-level phenomena in MT has led since 2010 to several initiatives and studies to improve the evaluation of pronoun translation, i.e.\ to make it more accurate but also more efficient, and if possible, to automate it.  With the advent of NMT, the new architectures have been submitted to the same tests and compared with PBSMT.  

ParCor is a parallel EN/DE corpus first annotated with anaphoric relations, and then with coreference ones \citep{GuillouEA:LREC14,lapshinova2018parcorfull}.  It includes TED talks and EU Bookshop publications.  The annotation pays special attention to the status of pronouns, and distinguishes several cases of referential vs.\ non referential uses.  Using similar annotation guidelines, the authors designed the PROTEST test suite, which contains 250 pronouns along with their reference translations \citep{GUILLOU16.327}.  Identity between a candidate and reference pronoun translation is scored automatically, but differences are submitted to human judgment.  Indeed, depending on the pronoun systems of the source and target languages (often EN/FR and EN/DE in these evaluations), but also and crucially on the lexical choice for a pronoun's antecedent, a variety of translations can be acceptable for pronouns.  This limits the accuracy of automatic reference-based metrics such as APT \citep{W17-4802}, as recently discussed by \citet{D18-1513}, and requires alternative strategies when evaluations must be quick, large-scale and cost-effective, e.g.\ for pronoun-oriented shared tasks. 



Several shared tasks have been organized to assess the quality of pronoun translation, but due to evaluation difficulties, protocols have evolved from year to year.  Two main approaches have been tried: (1)~evaluate the accuracy of pronoun \emph{translation}, though this cannot be done automatically with sufficient confidence for a shared task, and requires some form of human evaluation; (2)~evaluate the accuracy of pronoun \emph{prediction} given the source text and a lemmatized version of the reference translation with deleted pronouns, which can be done (semi-])automatically.\footnote{Lemmatization prevents non-MT strategies like powerful language models from attaining high scores, as it happened at the 2015 DiscoMT shared task \citep{W15-2501}.}  Both approaches have been tried at the DiscoMT 2015 shared task \citep{W15-2501}, but only the second one was continued in the following years \citep{W16-2345,W17-4801}.  


At WMT 2018, pronoun translation was evaluated for all 16 systems participating in the EN/DE news translation task using an external test suite \citep{W18-6435} in the PROTEST style, with 200 occurrences of \emph{it} and \emph{they} on the source side.  These pronouns have multiple possible translations into German.  Evaluation was semi-automatic, with candidates matching the reference (1,150) being `approved' and the others being submitted to human judges (2,050).  Seven out of 15 systems (all NMT) translate correctly more than 145 pronouns out of 200, with the best one reaching 157 \citep[Microsoft's Marian][]{P18-4020}.  Pronoun accuracy is highly correlated with BLEU ($r=$ 0.91) and APT ($r=$ 0.89).  
Event references reach 81\% accuracy and pleonastic \emph{it} 93\% on average. 
Intra-sentential anaphoric occurrences of \emph{it} are better translated than inter-sentential ones (58\% vs.\ 45\%).

A similar method using PROTEST was applied by the same authors to EN/FR MT with PBSMT and NMT systems \citep{hardmeier2018pronoun}, with 250 occurrences of \emph{it} and \emph{they}.  The best system, which is the Transformed-based context-aware system from \citet{P18-1117},
translated 199 pronouns correctly, while the average over the 9 tested systems is 160 (64\%).  The study shows that the context-aware system is highly accurate on pleonastic (non-referential) pronouns (27 out of 30) and intra-sentential anaphoric \emph{it} and \emph{they} (35/40 and 21/25) but still struggles with inter-sentential ones (15/30 and 11/25).


The evaluation approach adopted by \citet*{P18-1117} for their context-aware NMT architecture 
is quite exemplary.  Their goal is to demonstrate improvement of pronoun translation, and this is evaluated without the use of specific test suites or contrastive pairs (see below).  The authors use Stanford's CoreNLP coreference resolution system\footnote{An unspecified component of CoreNLP \citep{P14-5010}, possibly the deterministic one \citep{J13-4004}.}
to identify sentences with pronouns that have their antecedent in the previous sentence; for such sentences, BLEU improves more (+1.2) than on average.  Moreover, for sentences with \emph{it} and a feminine antecedent, BLEU increases by 4.8 points.  The attention weights of the system are compared to CoreNLP results in two ways, first by identifying the token that receives maximal attention when the pronoun is translated.  This token coincides with the antecedent found by CoreNLP more often (+6 points) than for baseline methods (random, first, or last noun).  The second evaluation has human raters identify the actual antecedent in 500 sentences with \emph{it} where more than one candidate antecedent exists in the previous sentence.  Here, CoreNLP is correct in only 79\% of the cases, while using NMT attention is 72\% correct, well above the best heuristic  at 54\%.  These arguments strongly indicate that NMT learns to perform inter-sentential anaphora resolution to some extent.\footnote{This analysis of performance in situations that are genuinely ambiguous is reminiscent of \citet*{Q16-1037}, who assess the capacity of LSTM networks to model syntactic dependencies such as noun-verb agreement.}

Moving away from evaluations performed on test suites with reference translations, as well as from those requiring coreference resolution, contrastive pairs have also been designed for pronoun translation.  The above-mentioned test set by \citet*{N18-1118} also contains 100 blocks 
that
aim to test the translation of personal and possessive pronouns.  As for WSD, the context and source sentences are kept constant (e.g.\ with a pronoun \emph{it}), but four alternative translations of the context sentence are generated, varying the translation of the antecedent: (a)~reference translation; (b)~possible translation with the opposite gender; (c)~and (d), inaccurate translations, feminine and masculine.  For each translation of the context sentence, the contrastive pair differs only in the translation of \emph{it}, with a masculine vs.\ feminine French pronoun (\emph{il} or \emph{elle}).  In situations (c) and (d), the system is expected to prefer the ``contextually correct'' translation, agreeing with the gender of the inaccurate translation of the antecedent.  The best system designed by \citet{N18-1118} achieves 72.5\% accuracy versus 50\% for a non-contextual NMT system.


Finally, a much larger but less structured set of contrastive pairs for pronouns has been presented by \citet*{W18-6307}.\footnote{See \url{https://github.com/ZurichNLP/ContraPro}.}  The EN/DE pairs contain only source sentences occurring in the Open Subtitles, without editing of the context sentences.  The key to ensure high quality automatic data selection is to focus on the English source pronoun \emph{it} and its possible German translations into \emph{er}, \emph{sie} or \emph{es}, with several constraints: automatic anaphora resolution on both EN/DE sides (with CoreNLP \citep{P14-5010}\footnote{Unspecified coreference component.} and CorZu \citep{Tuggener2016Thesis}) must find an antecedent; the antecedents on the EN and DE sides must be word aligned (with fast-align \citep{N13-1073}); and the EN/DE pronouns must also be aligned.  With these constraints in mind, the set includes the source and target sentences containing \emph{it} and its translation, and as much context as needed before the sentence.  To generate the wrong alternative in the contrastive pair, the correct translation is randomly replaced with one of the two incorrect ones.  The set contains 12,000 occurrences of \emph{it}, with 4,000 for each possible translation; most antecedents (58\%) are in the previous sentence.
Using these contrastive pairs, the authors find that context-aware models outperform baselines by up to 20 percentage points,  especially on the sentences where the antecedent is in the preceding sentence, 
while BLEU scores are only marginally improved.  



\subsubsection{Evaluation of Discourse Structure}
 
Several metrics have been proposed to assess the ability to correctly translate discourse structure, but none of the studies applied them to NMT systems, as they pre-dated their advent.  Discourse structure results from argumentation relations between sentences, often made explicit through discourse connectives.  Although strategies to improve connective translation by PBSMT systems have been designed \citep{W12-0117,meyer2015disambiguating}, along with metrics to assess the improvements \citep{hajlaoui2013assessing}, they have not been recently applied to NMT systems.  

Similarly, metrics involving discourse structure (sentence-level RST parse trees) such as DiscoTKparty \citep{J17-4001} have been shown to correlate positively with human judgments (for PBSMT). However, this study mostly refers to data from the WMT 2014 shared task on (meta-)evaluation of metrics, which did not include any NMT output at that time. 
\citet{smith2018assessing} designed a discourse-aware MT evaluation metric that compares embeddings of source and target connectives, which is validated on EN/FR MT outputs from 2014 and earlier, accompanied by human ratings.  

A manual analysis of discourse phenomena in SMT, with quality estimation as the background objective, was presented by \citet*{Scarton2015Quantitative}, while other taxonomies of discourse-related errors, applied by manual analysts, have been inspired by contrastive linguistics at the discourse level, allowing comparison of cross-lingual contrasts in human and machine translation and concluding to NMT superiority \citep{W17-4810,W18-6305}.

Indirect evidence on the capability of NMT to translate inter-sentential dependencies comes from the recent study by \citet*{D18-1512}, reassessing \citeauthor{Hassan2018AchievingHumanParity}'s claim that the Bing Translator achieved human parity on ZH/EN news translation.  Without examining detailed quality attributes such as word order, lexical choice, or pronouns, the authors asked human judges to rate translations at the text level rather than the sentence level, and they showed that when entire texts are considered by professional translators, the difference between human and NMT translations becomes statistically significant.  One can therefore infer that there are perceptible imperfections in the NMT translation of text-level properties such as cohesion and coherence.


\subsection{Synthesis} 

When a new system is presented in a publication, it cannot be expected from the authors that they apply a large array of existing metrics.  Evaluation studies that deepen the analysis of MT output are thus welcome.  As reviewed in this section, studies of NMT models from 2017--2018 have revealed significant improvements in output quality brought by NMT models, confirmed from a variety of perspectives:
\begin{itemize}
	\item type of metric: automated (e.g.\ using the TER distance to a reference translation) vs.\ human (e.g.\ judges who may post-edit, or compare, or rate absolutely one or more translations, with or without knowledge of the source language);
	\item type of comparison: absolute score or comparative score (often pitching NMT against SMT);
	\item type of system output: 1-best, $n$-best, or probabilities over contrastive pairs (which require access to a system's internals);
	\item test data: large corpora from WMT, excerpts from them, domain-specific data, or test suites aimed at one or more linguistic phenomena.
\end{itemize}
One of the main observations of this review is the rather large number of assessments of document-level quality, which frequently support the need for discourse-aware MT.  However, these studies also indicate that significant progress remains to be made on several discourse-level phenomena: lexical coherence, anaphora resolution. and discourse structure.  


%% file: context-nmt.tex
\section{Increasing Context Spans in NMT} 
\label{sec:dlnmt}

In a subsequent version of the paper, this section will review studies from 2017--2018 that attempted to improve the translation of discourse-level phenomena, and/or attempted to use larger spans of context when translating.  The section will be divided in three parts: NMT systems using wider contexts; NMT models for improving WSD and lexical choice; and the processing of discourse-level phenomena, particularly pronouns, in NMT.